\documentclass[10pt,twocolumn,letterpaper]{article}

\usepackage[pagenumbers]{cvpr} % To force page numbers, e.g. for an arXiv version

\usepackage[dvipsnames]{xcolor}

\usepackage{multirow}
\usepackage{makecell}
\usepackage[symbol]{footmisc}
\definecolor{cvprblue}{rgb}{0.21,0.49,0.74}
\usepackage[pagebackref,breaklinks,colorlinks,citecolor=cvprblue]{hyperref}

\def\paperID{3869} % *** Enter the Paper ID here
\def\confName{CVPR}
\def\confYear{2024}

\title{ChartLlama: A Multimodal LLM for Chart Understanding and Generation}

\author{Yucheng Han$^{1,2\ast}$ 
\quad 
Chi Zhang$^{2\ast\dagger}$ 
\quad Xin Chen$^{2}$ \quad Xu Yang$^{3}$ \\ \quad Zhibin Wang$^2$ \quad Gang Yu$^{2}$ \quad Bin Fu$^2$ \quad Hanwang Zhang$^1$ \vspace{0.3em} \\
{\normalsize $^1$Nanyang Technological University} \quad
{\normalsize $^2$ Tencent} \quad
{\normalsize $^3$ Southeast University} \\
{\normalsize $^1$ \{yucheng002,~hanwangzhang\}@ntu.edu.sg}  \\
{\normalsize $^2$ \{johnczhang,~shingxchen,~billzbwang,~skicyyu,~brianfu\}@tencent.com} \quad {\normalsize $^3$ xuyang\_palm@seu.edu.cn} \\
\url{https://tingxueronghua.github.io/ChartLlama/}
}

\begin{document}
\twocolumn[{
\renewcommand\twocolumn[1][]{#1}
\maketitle
\centering
\vspace{-0.5cm}
\includegraphics[width=.99\linewidth]{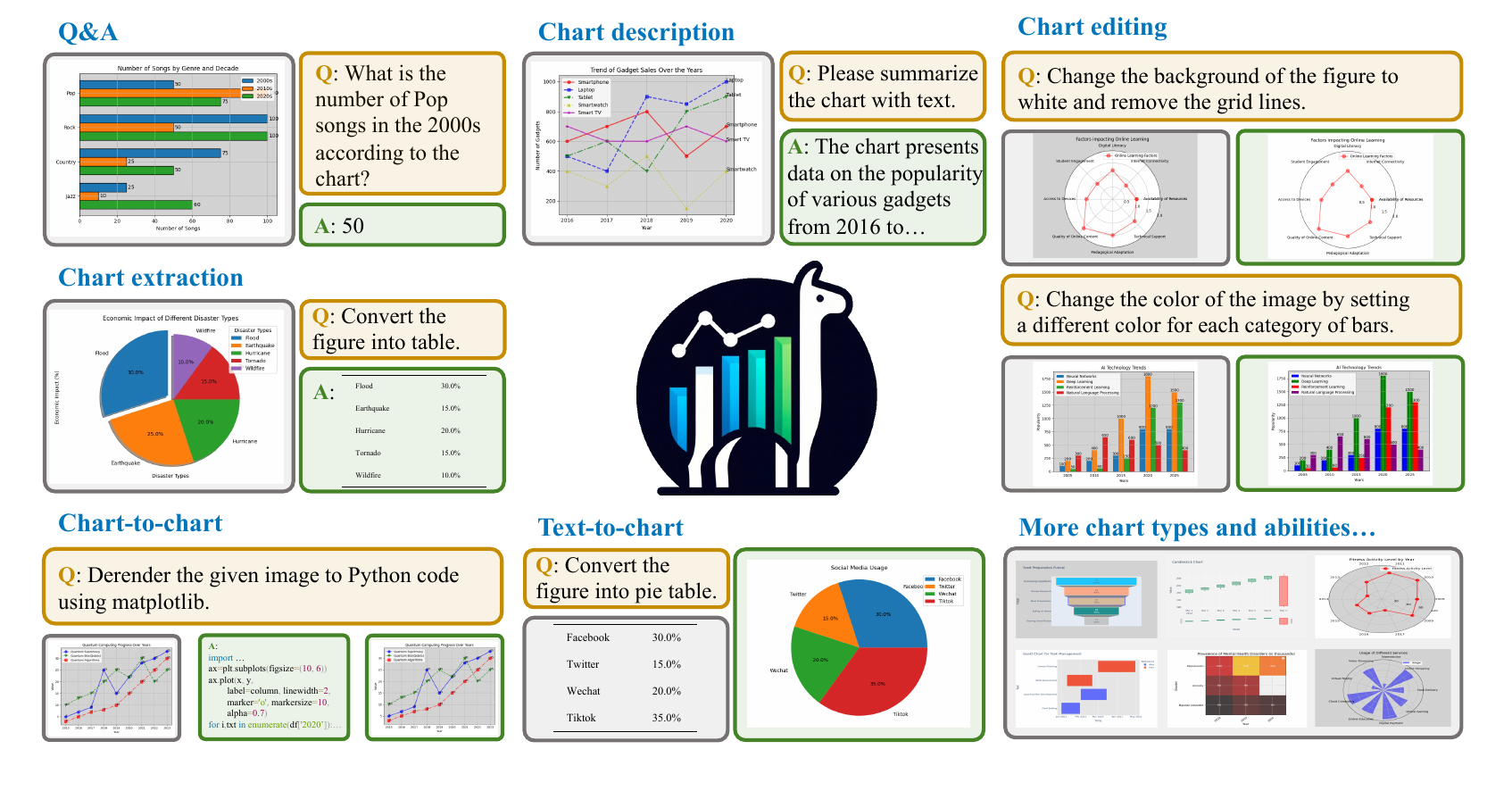}
\captionsetup{type=figure}
\caption{\textbf{Capability demonstration of ChartLlama.} An instruction-tuning dataset is created based on our proposed data generation pipeline. We train ChartLlama on this dataset and achieve the abilities shown in the figure.}
\vspace{0.2cm}
\label{fig:teaser}
}]

\footnotetext[1]{Equal contributions. Work was done when Yucheng Han was a Research Intern at Tencent.}
\footnotetext[2]{Corresponding Author.}

\begin{abstract}
Multi-modal large language models have demonstrated impressive performances on most vision-language tasks. However, the model generally lacks the understanding capabilities for specific domain data, particularly when it comes to interpreting chart figures. This is mainly due to the lack of relevant multi-modal instruction tuning datasets. 
In this article, we create a high-quality instruction-tuning dataset leveraging GPT-4. We develop a multi-step data generation process in which different steps are responsible for generating tabular data, creating chart figures, and designing instruction tuning data separately. 
Our method's flexibility enables us to generate diverse, high-quality instruction-tuning data consistently and efficiently while maintaining a low resource expenditure. Additionally, it allows us to incorporate a wider variety of chart and task types not yet featured in existing datasets.
Next, we introduce ChartLlama, a multi-modal large language model that we've trained 
% fine-tuned
using our created dataset. ChartLlama outperforms all prior methods in ChartQA, Chart-to-text, and Chart-extraction evaluation benchmarks. Additionally, ChartLlama significantly improves upon the baseline in our specially compiled chart dataset, which includes new chart and task types. The results of ChartLlama confirm the value and huge potential of our proposed data generation method in enhancing chart comprehension. 
\end{abstract}    
\section{Introduction}
\label{sec:intro}

In the past year, the field of artificial intelligence has undergone remarkable advancements.  A key highlight is the emergence of large language models (LLMs) like GPT-4~\cite{gpt4}. These models~\cite{instructgpt,glm,2023internlm,baichuan2023baichuan2,llama,llama2} have demonstrated a remarkable capability to comprehend and generate intricate textual data, opening doors to myriads of applications in both academia and industry.
Taking this progress a step further, the introduction of GPT-4V~\cite{gpt4v} marked another milestone. It endows LLMs with the ability to interpret visual information, essentially providing them with a vision.
As a result, they can now extract and analyze data from images, marking a significant evolution in the capacities of these models.

However, despite the achievements and potentials of models like GPT-4V, the details behind GPT-4V's architecture remain a mystery.
This opacity has given rise to questions within the academic world about the best practices for designing multi-modal LLMs.
Notably, pioneering research initiatives, like LLaVA~\cite{llava,llava1.5} and MiniGPT~\cite{minigpt,minigptv2}, provide insightful directions in this regard.
Their findings suggest that by incorporating visual encoders into existing LLMs and then 
fine-tuning them using multi-modal instruction-tuning datasets, LLMs can be effectively transformed into multi-modal LLMs.
It's noteworthy that these multi-modal datasets are typically derived from established benchmarks, presenting a cost-effective method for accumulating data required for instruction tuning.

Datasets grounded on established benchmarks, such as COCO~\cite{coco}, have significantly enhanced the abilities of multi-modal LLMs to interpret everyday photographs adeptly.
However, when confronted with specialized visual representations, such as charts, they reveal a noticeable limitation~\cite{gpt4v, hallusionbench}.
Charts are important visual instruments that translate complex data sets into digestible visual narratives, playing a crucial role in facilitating understanding, shaping insights, and efficiently conveying information.
Their pervasive presence, from academic publications to corporate presentations, underscores the essentiality of enhancing the capability of multi-modal LLMs in interpreting charts.
Indeed, gathering data specifically to refine instructions for understanding charts presents several challenges. These typically stem from two areas: understanding and generation. An effective chart understanding model should be capable of extracting and summarizing data from various types of charts and making predictions based on this information. 

However, most existing datasets~\cite{chartqa, chart-to-text, plotqa, unichart} only provide support for simple question-answering or captioning, primarily due to the absence of detailed chart information and annotations that provide a high-level understanding of raw data. The high dependency on manually annotated charts gathered by web crawlers negatively affects the quality of these datasets. Thus, the previous annotating methods could only result in chart datasets with lower quality and less comprehensive annotations.
Compared with chart understanding, generating chart figures is a more challenging task for the model because existing
deep-learning-based generation methods~\cite{dalle,stablediffusion} struggle to accurately create images based on instructions. 
Using Python code to generate charts seems promising which needs the corresponding annotations to supervise models. Most charts obtained from the web are devoid of detailed annotations, making it challenging to annotate the generation code. The absence of code annotations makes it challenging to supervise models in code generation. These issues combined impede the model's ability to understand charts and learn generation jointly.

To address this,
we introduce an adaptive and innovative data collection approach exclusively tailored to chart understanding and generation. 
At the heart of our methodology is the strategic employment of GPT-4's robust linguistic and coding capabilities, which facilitate the creation of rich multi-modal datasets. This innovative integration not only optimizes data accuracy but also ensures its wide-ranging diversity.
Specifically, our method comprises three main phases:
\\1) \textbf{Chart Data Generation}.
Our strategy for data collection stands out for its flexibility. Rather than limiting data collection to conventional data sources such as the web or existing datasets, we harness the power of GPT-4 to produce synthesized data. By providing specific characteristics such as topics, distributions, and trends, we guide GPT-4 to produce data that is both diverse and precise.
\\2) \textbf{Chart Figure Generation}.
Subsequently, GPT-4's commendable coding skills are utilized to script chart plots using the open-sourced library, like Matplotlib, given the data and function documentation. 
The result is a collection of meticulously rendered charts that span various forms, each accurately representing its underlying data.
\\3) \textbf{Instruction data generation}.
Beyond chart rendering, GPT-4 is further employed to interpret and narrate chart content, ensuring a holistic understanding. It is prompted to construct relevant question-answer pairs correlating with the charts. 
This results in a comprehensive instruction-tuning corpus, amalgamating the narrative texts, question-answer pairs, and source or modified codes of the charts.

A standout feature of our methodology is its flexibility, which diminishes the potential for bias while simultaneously offering scalability. 
Building on this robust methodology, we've crafted a benchmark dataset, which is made available for public access. 
This dataset stands out, not only for its superior quality but also its unparalleled diversity. A comparative analysis of our benchmark against existing datasets can be viewed in Table~\ref{tab:dataset_statistics}.
To showcase the superiority of our benchmark, we introduced a multi-modal Large Language Model (LLM) named ChartLlama trained with our established benchmarks. 
Our extensive experiments evaluated on multiple existing benchmark datasets show that our model outperforms previous methods with remarkable advantages and considerably less training data.
Additionally, ChartLlama is equipped with several unique capabilities, including the ability to support a wider range of chart types, infer across multiple charts, undertake chart de-rendering tasks, and even edit chart figures.

Our main contributions are summarized as follows:
 \begin{itemize}
	\item We introduce a novel multi-modal data collection approach specifically designed for chart understanding and generation. The proposed data collection method boasts superior flexibility and scalability, enabling easy migration to different types of charts and various tasks.
	\item Through our innovative data collection approach, we create a benchmark dataset that stands out in terms of both quality and diversity.
 We make this dataset publicly available to catalyze further advancements in the field.
	\item We develop ChartLlama, a multi-modal LLM that not only surpasses existing models on various existing benchmarks but also possesses a diverse range of unique chart understanding and generation capabilities.

\end{itemize}
\section{Related work}
\label{sec:related_work}

\begin{table}[t]
\resizebox{0.45\textwidth}{!}{
\begin{tabular}{ccccc}\toprule[1pt]
Datasets & \#Chart type & \#Chart figure & \begin{tabular}[c]{@{}c@{}}\#Instruction \\ tuning data\end{tabular} & \#Task type \\ \midrule[1pt]
ChartQA~\cite{chartqa} & 3 & 21.9K & 32.7K & 1 \\
PlotQA~\cite{plotqa} & 3 & 224K & 28M & 1 \\
Chart-to-text~\cite{chart-to-text} & 6 & 44K & 44K & 1 \\
Unichart~\cite{unichart} & 3 & 627K & 7M & 3 \\
StructChart~\cite{structchart} & 3 & 9K & 9K & 1 \\
\textbf{ChartLlama} & \textbf{10} & \textbf{11K} & \textbf{160K} & \textbf{7} \\ \bottomrule[1pt]
\end{tabular}
}
\caption{\textbf{Dataset statistics}. Thanks to the flexibility of our data construction method, our proposed dataset supports a wider range of chart types and tasks. We can generate more diverse instruction-tuning data based on specific requirements. }
\label{tab:dataset_statistics}
\end{table}
\begin{figure}[t]
    \centering
    \includegraphics[width=0.45\textwidth]{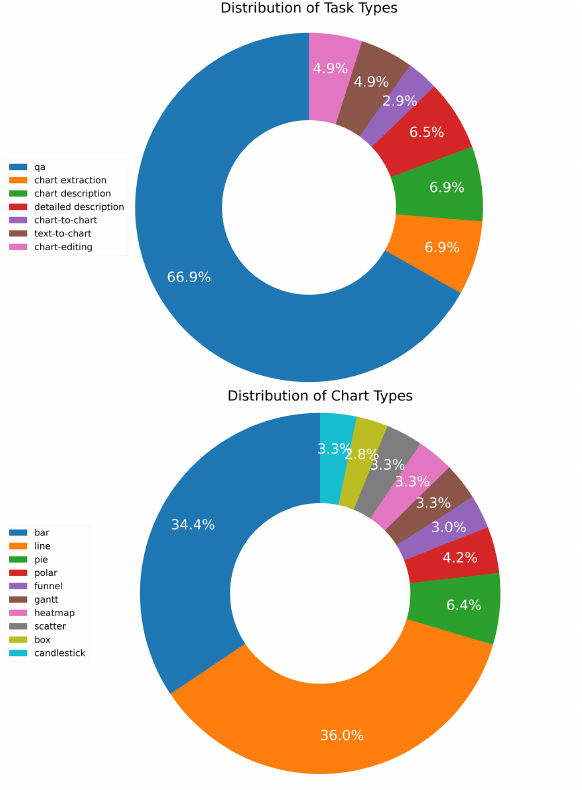}
    \caption{\textbf{Distributions of different types of data in our dataset}. The top and bottom pie charts show the distribution of task types and chart types, respectively. (The illustration is generated by our proposed ChartLlama.)}
    \label{fig:statistics}
\end{figure}

\begin{figure*}[t]
    \centering
    \includegraphics[width=\textwidth]{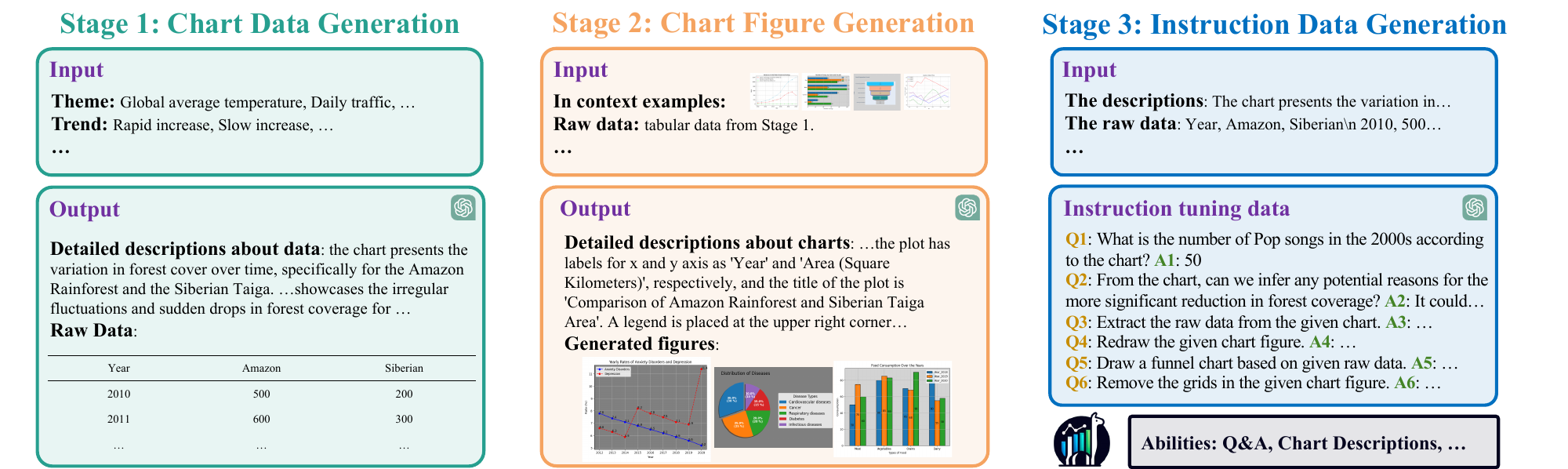}
    \caption{\textbf{Pipeline of our data generation method}. The innovative data generation process we proposed consists of three important steps relying on GPT-4. The dataset generated using this process exhibits significant advantages compared to previous datasets in terms of data diversity, quality, the number of chart types, and the variety of tasks. ChartLlama, which is trained on this dataset, has the ability to perform various tasks based on the design of the instruction-tuning data.}
    \label{fig:pipeline}
\end{figure*}
\subsection{Large Language Model}
The series of LLM models, such as GPT-3.5~\cite{instructgpt} and GPT-4~\cite{gpt4}, have demonstrated remarkable reasoning and conversational capabilities, which have garnered widespread attention in the academic community. Following closely, a number of open-source LLM~\cite{baichuan2023baichuan2,llama,llama2,glm,qwen} models emerged, among which Llama~\cite{llama} and Llama 2~\cite{llama2} are notable representatives. With extensive pre-training on large-scale datasets and carefully designed instruction datasets, these models have also showcased similar understanding and conversational abilities. Subsequently, a series of works have been developed, aiming to achieve specific functionalities by leveraging efficient supervised fine-tuning algorithms based on the Llama series. Among these influential works, Alpaca~\cite{alpaca} and Vicuna~\cite{vicuna} stand out, with Vicuna's framework serving as the cornerstone for subsequent multi-modal works.

\subsection{Multi-modal Large Language Model}
Concurrently, the academic community has witnessed a surge of development in multi-modal LLMs~\cite{li2023otter,ye2023mplugowl,li2023blip2,li2023finetuning,llamaadapter,bliva,mmicl,Qwen-VL,internlmxcomposer,llava1.5,llava,minigptv2,minigpt} built upon existing open-source models. Earlier efforts in this domain, such as LLaVA~\cite{llava}, MiniGPT~\cite{minigptv2}, BLIP2~\cite{li2023blip2}, and mPLUG-Owl~\cite{ye2023mplugowl}, have shown significant room for improvement in both performance and functionality. With further exploration of training strategies and an increase in dataset scale, the performance of these new models has steadily improved, reaching comparable levels to GPT-4V in specific evaluation metrics. Notably, LLaVA-1.5~\cite{llava1.5}, an iterative version of LLaVA, has gained popularity as a baseline due to its user-friendly training framework, superior performance, and data efficiency. Our work is also based on LLaVA-1.5.

\subsection{Chart Understanding}
In evaluations such as the report of GPT-4V~\cite{gpt4v} and HallusionBench~\cite{hallusionbench}, it is evident that current multi-modal LLMs still struggle with complex chart-related problems. There are already some datasets~\cite{plotqa,chart-to-text,chartqa} available for evaluating models' chart understanding capabilities, mainly divided into two categories, each with its own advantages and disadvantages. One category measures through simple question-and-answer tasks, such as ChartQA~\cite{chartqa}, which has high-quality questions and answers annotated by humans, and PlotQA~\cite{plotqa}, which has lower-quality questions and answers generated through templates. The advantage of these datasets lies in their large scales and the ability to generate them through templates. However, their limitations include the difficulty in ensuring the quality of questions and answers, as well as a tendency to focus too much on simple questions about the data in the charts. The other category converts charts into textual descriptions, with Chart-to-text~\cite{chart-to-text} being a representative work in this field. The charts and annotations in these datasets are derived from the real world, ensuring higher quality, and encouraging models to delve deeper into the trends and meanings behind the charts. However, the corresponding drawbacks include the presence of more noises in the textual annotations and the over-reliance on BLEU-4. Previous works focusing on chart understanding tasks can be divided into two main kinds of approaches. One kind of approach is using a single model to understand the charts and answer questions in natural language, for example,~\cite {unichart,matcha}. The other kind of approach, such as~\cite{liu2022deplot,structchart}, is to first utilize the model to convert the charts into structured data and then analyze and answer questions based on the structured data using existing large models. In our work, we primarily explore the former kind, aiming to leverage a single model to complete the entire process of chart understanding.

\section{Method} 
\label{sec:method}

In this section, we detail our unique approach to chart understanding and generation. Our method involves three interconnected steps: data collection, chart figure generation, and instruction data generation. We illustrate this process in Fig.~\ref{fig:pipeline}. These steps are detailed in the following subsections.

\subsection{Chart Data Generation}
Our primary goal in chart data collection is to collect diverse and high-quality data. We employ two main strategies for this purpose: 1) Data Generation from Scratch Using GPT-4: To collect a diverse and high-quality dataset, we initially generate tabular data from scratch using GPT-4. We instruct GPT-4 to create data tables based on specific themes, distributions, and other characteristics like the size of the dataset in terms of rows and columns.
This process ensured the creation of data with known and controlled characteristics, which can be essential for generating reliable instruction-answer pairs. Moreover, by managing these characteristics, we can intentionally minimize bias, leading to a more balanced dataset.
2) Synthesizing Data from Existing Chart Datasets. Our second strategy is to synthesize data by referencing existing chart datasets. These datasets already encompass a range of topics and characteristics, providing a solid base for data generation. By prompting GPT-4 with these datasets, we guide it to generate reasonable data that complements its existing knowledge base. This method added variety to our dataset and improved its overall quality.

Generating
diverse data at scale using the
LLM is not an easy task. When the prompt is designed improperly, the model tends to generate repetitive and meaningless data that deviates from
the distribution of real-world data and thus lacks valuable insights that could be important for designing meaningful question-and-answer tasks. If we simply provide a set of data and require the model to imitate without any additional guidance, the model will probably just repeat the reference data. Therefore, in this step, it is necessary to provide the model with additional information, such as the topic and distribution, to ensure that it can be properly guided to generate meaningful data. We will now explain these pieces of information in detail.

\noindent\textbf{Chart theme:} We first generate hundreds of possible themes, which are all short phrases. 
When we 
generate data, we randomly select one from all those themes, which makes the data meaningful and diverse.  This also makes it much more easy to generate questions and responses for instruction tuning. 

\noindent\textbf{Data trends:} Another important characteristic of the data is the trends. We first generate several typical trend descriptions, like steadily increasing and suddenly dropping, then randomly select a few trends and require the model to generate data following them. If lacking such characteristics, the model will tend to generate several sets of data with meaningless distributions. 

\noindent\textbf{Column and row lengths:} The lengths of columns and lengths are also necessary for data generation. Without specific constraints, LLMs tend to generate excessively long or even repetitive data, which is difficult to present in a meaningful way through charts.

\noindent\textbf{Chart types:} Charts of different types usually share different characteristics. For example, the sum of the values in pie charts should be 100\%. If not specify the type of chart, we might end up generating data that doesn't comply with the corresponding chart standards.

\subsection{Chart Figure Generation}
The next step is to transform our dataset into visual charts using GPT-4's coding capabilities. We used popular chart plotting libraries, such as Matplotlib, as our primary tools. When prompting GPT-4, we provide the collected data, relevant function documentation, and in-context examples. We also give detailed instructions on diversifying aspects like color schemes and line types to enhance the visual appeal of the charts.
To increase the diversity and success rate of our chart generation, we randomly sample successfully generated codes as in-context examples in the prompts.
Compared with previous automated chart generation efforts that relied on templates, our approach offers greater variety and better visual appeal. It also enables us to generalize across different chart types effectively. 
The result was a collection of meticulously crafted charts, each accurately representing its data and visually appealing, showcasing the effectiveness of our method. The necessary input for the prompts in this stage is listed below. 

\noindent\textbf{Chart data:} This is the most essential input for the task. The chart data is the information that will be visualized in the chart. Without it, no meaningful chart can be made.

\noindent\textbf{Related function documentation:} This is an important reference for generating the Python code. It provides information about the available functions and features that can be used to create the chart. With the documentation, the model could even create charts in new styles that are not in the in-context examples.

\noindent\textbf{In context example:} These in-context examples are sampled from pre-selected high-quality code. This helps to facilitate the construction of the Python code. When there is new generated code in high quality, we can save and sample it, which is used as in-context examples later.

\noindent\textbf{Other requirements:} To ensure that the final generated code is suitable for batch processing and execution, we also need to include several requirements in the prompt. For example, the data is required to be listed in the code to make the generated code self-contained and executable without the need for external files. We also set the requirements for the title, axis labels, legend, and text annotations. They provide context about what the chart represents and make it easier to understand the data. Without them, the chart can be confusing and difficult to interpret.

\subsection{Instruction data generation}
After completing the first two stages, we gathered comprehensive information about each chart, including precise tabular data, various characteristics from various perspectives, and the chart plotting code.
Leveraging this rich information, we move on to generating a wide range of instruction-answer data with the assistance of GPT-4, significantly enhancing the capabilities of models trained on this dataset. In addition to fundamental chart understanding functionalities such as Q\&A and summarization, our approach allows us to construct instructions and answers for more complex tasks, such as accurate data extraction, detailed chart descriptions, chart code generation, and even chart editing.
Compared to previous pipelines for instruction data generation that often rely on human annotation, our methods yield significant time savings while enhancing diversity and quality in the resulting dataset.

Here are more details about the data that needs to be filled into the prompt.

\noindent\textbf{Chart descriptions and raw data:} providing these descriptions helps the model understand the context better. The first description helps the model to understand the nature of the data, and the second description assists in understanding the visual representation of the data. The raw data feeds the model with the actual values to base its responses on. All the descriptions and raw data are generated in the first and second stages.

\noindent\textbf{Characteristics to be asked about:} This requirement ensures that the model asks diverse and relevant questions about the chart. It prompts the model to explore different features of the data and its representation.
\section{Experiment}

\begin{table*}[t!]
\resizebox{0.95\textwidth}{!}{
\begin{tabular}{l|ccc|cc|cc|cc} \Xcline{1-10}{1pt}
\multirow{2}{*}{Method} & \multicolumn{3}{c}{Chartqa} & \multicolumn{2}{c}{Chart-to-text} & \multicolumn{2}{c}{Chart extraction (human)} & \multicolumn{2}{c}{Chart extraction (augmented)} \\ \Xcline{2-10}{1pt}
 & Human & Augmented & Average & Pew & Statista & Precision & F1 & precision & F1 \\ \Xcline{1-10}{1pt}
Pix2struct~\cite{lee2023pix2struct} & 30.50 & 81.60 & 56.00 & 10.30 & 38.00 & -- & -- & -- & -- \\
Matcha~\cite{matcha} & 38.20 & 90.20 & 64.20 & 12.20 & 39.40 & -- & -- & -- & -- \\
DePlot~\cite{liu2022deplot} & -- & -- & -- & -- & -- & 81.32 & 81.15 & 93.42 & 93.29 \\
Unichart~\cite{unichart} & 43.92 & 88.56 & 66.24 & 12.48 & 38.21 & 61.51 & 35.20 & 79.59 & 70.21 \\
Baseline*~\cite{llava1.5} & 37.68 & 72.96 & 55.32 & 7.16 & 24.65 & 53.48 & 48.39 & 55.17 & 49.50 \\
\textbf{ChartLlama} & \textbf{48.96} & \textbf{90.36} & \textbf{69.66} & \textbf{14.23} & \textbf{40.71} & \textbf{84.92} & \textbf{84.89} & \textbf{94.94} & \textbf{94.78} \\  \Xcline{1-10}{1pt}
\end{tabular}
}
\caption{\textbf{Results on traditional tasks.} We compare our work with the previous three open-source models and also compare it with Baseline* trained on the training split of respective benchmarks. }
\label{tab:main_result}
\end{table*}
\begin{table*}[t]
\resizebox{0.95\textwidth}{!}{
\begin{tabular}{c|c|cc|cc|cc|cc}\Xcline{1-10}{1pt}
\multirow{2}{*}{Method} & \multirow{2}{*}{Detailed Description} & \multicolumn{2}{c|}{Chart-to-chart} & \multicolumn{2}{c|}{Text-to-chart} & \multicolumn{2}{c|}{Chart-editing} & \multicolumn{2}{c}{Chart-to-text} \\ \Xcline{3-10}{1pt}
 &  & GPT Score & Success Rate (\%) & GPT Score & Success Rate (\%) & GPT Score & Success Rate (\%) & Pew & Statista \\ \Xcline{1-10}{1pt}
LLaVA-1.5~\cite{llava1.5} & 67.2 & 64.8 & 46 & 62.2 & 77 & 51.6 & 38 & 65.8 & 73.4 \\
\textbf{ChartLlama} & \textbf{74.2} & \textbf{74.4} & \textbf{73} & \textbf{81.6} & \textbf{81} & \textbf{75.6} & \textbf{71} & \textbf{81.0} & \textbf{92.6} \\ \Xcline{1-10}{1pt}
\end{tabular}
}
\caption{\textbf{Results on new tasks.} We primarily compared our work with the baseline model LLaVA-1.5. For the proposed new task, we used GPT for evaluation and validated the effectiveness of our proposed dataset. Evaluation of Chart-to-text using ChatGPT is also listed.}
\label{tab:gpt_eval}
\end{table*}
\begin{table}[t!]
\resizebox{0.45\textwidth}{!}{
\begin{tabular}{cccc}\toprule
Chart type  & Unichart    & Baseline* & \textbf{ChartLlama}        \\ \midrule
Funnel      & 18.30       & 49.32     & \textbf{70.59}      \\
Gantt       & 9.80       & 40.17     & \textbf{56.64}      \\
Heatmap     & 25.43      & 38.18     & \textbf{53.18}      \\
Scatter     & 26.32      & 37.91     & \textbf{54.97}      \\
Box         & 16.67      & 28.33     & \textbf{37.33}      \\
Candlestick & 15.79      & 25.69     & \textbf{46.20}       \\ \bottomrule
\end{tabular}
}
\caption{\textbf{Performances of Q\&A on more categories of chart.} Baseline* means a modified version of LLaVA-1.5, which is further trained on the ChartQA dataset. We evaluate the performance of Baseline* and the previous state-of-the-art model Unichart on these new chart types.}
\label{tab:special_chart_results}
\end{table}
\subsection{Implementation details and dataset statistics}
\noindent\textbf{Implementation details.} We train ChartLlama based on LLaVA-1.5 which provides fundamental abilities crucial for chart understanding and generation, including the OCR functionality. The projection layer and LLM are trained on our proposed dataset. Details of the model architecture and training hyper-parameters can be referred to in our appendix. 

\noindent\textbf{Dataset statistics.} We show the statistics of our generated dataset in Table~\ref{tab:dataset_statistics} and Figure~\ref{fig:statistics}.
In our instruction-tuning data, Q\&A dominates while the other tasks correspond to similar proportions of data. This is mainly because a single chart could be utilized to construct multiple Q\&A data. 
Previous datasets usually gather only three types of charts: bar charts, line charts, and pie charts. Unlike them, we support a wide range of chart types. This is mainly due to the strong flexibility of our data construction method. It's worth noting that we can continue to expand on more data and chart types in the future. 

\subsection{Evaluation Benchmark and Metrics}\label{sec:tasks_and_evaluation}
We evaluate possible models on seven tasks, including both the traditional tasks and novel tasks which verifies that our data generation pipeline has good scalability towards various tasks and chart types. 

\noindent\textbf{Traditional Tasks.} 
Three traditional tasks are evaluated, namely ChartQA, Chart-to-text, and Chart-extraction.
\\1) For ChartQA~\cite{chartqa}, we evaluate relaxed accuracy on human and augmentation split, respectively. The question-and-answer data on the human split is more challenging because it includes more questions that require mathematical reasoning.
\\2) Chart-to-text contains two separate datasets for training and evaluation. BLEU-4 and GPT-4 serve as metrics for evaluation. BLEU-4 is widely used in many NLP tasks. However, when it comes to Chart-to-text, there's a critical issue. The Chart-to-text datasets contain too few ground-truth references, which means that the results must be very close to the reference targets to achieve high scores. Thus, we have to train ChartLlama on the train split when evaluating using BLEU-4. To facilitate more reasonable evaluations, we propose a new evaluation metric based on GPT-4, referring to the GPTScore~\cite{gptscore}. We designed scoring criteria that require the ground-truth reference and raw data as input conditions. Details can be found in the appendix.
\\3) Chart extraction aims to extract the tabular data from the given chart figure. We follow the evaluation framework of DePlot~\cite{liu2022deplot} and report the Precision and F1 scores on the challenging ChartQA dataset, which also provides the tabular data for each chart figure.

\noindent\textbf{New tasks.}
In addition to traditional tasks, we have devised four additional innovative tasks, three of which are targeted at chart generation to verify the scalability of novel tasks. 
\\1) Detailed description. This task necessitates a comprehensive description of the given chart figure in a detailed manner, rather than summarizing it briefly. The evaluation metric for detailed description is similar to the evaluation metric in Chart-to-text using GPT-4. We include detailed descriptions of the data and chart figures as conditions for GPT-4 to assist evaluation. Additionally, our evaluation criteria are more exhaustive, outlining various elements that the model under evaluation should generate. These elements include the data characteristics and visual attributes of the chart figures. 
\\2) Chart-to-chart. This task aims to reconstruct the given chart figure. We design comprehensive evaluation metrics for code generation and utilize GPT-4 to measure the quality of the code. For the chart-to-chart task, we evaluated the precision of data, axes, colors, chart types, and titles, rating from 0 to 5. Then we average them as the score for each sample. Finally, we normalize it to a range of 0 to 100 for easier analysis and report the average score across the entire test set.
\\3) Text-to-chart. The task aims at generating chart figures according to instructions and tabular data. We provide the input instructions and the generated code as conditions for evaluation criteria. The evaluation focuses mainly on visual similarity, completeness, accuracy, and aesthetics. Each standard is equally rated from 1 to 5 points. After averaging and normalization, we get the final score. 
\\4) Chart-editing. The input condition for this task is a chart figure and an instruction describing how to edit the chart. It is expected to create a new figure that has been modified according to instructions based on the given chart figure. The evaluation method for chart-editing uses a similar process to previous chart generation-related tasks. The input conditions include the code of the chart to be modified, instructions, and the generated code of the model. The data accuracy, completeness, aesthetics, and instruction following performance are scored on a scale from 0 to 5. After averaging and normalization, the final result is obtained.
\\For further details, please refer to the appendix.

\begin{figure*}[t]
    \centering
    \includegraphics[width=0.9\textwidth]{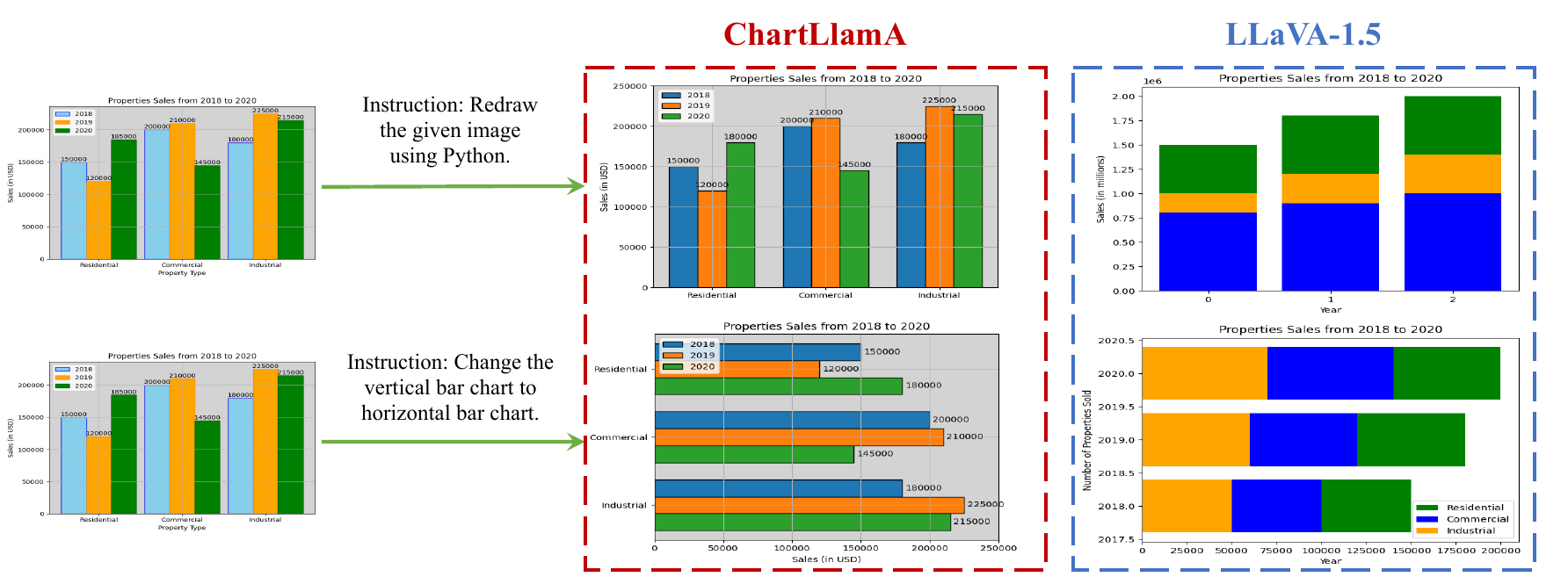}
    \caption{\textbf{Qualitative comparison for Chart-to-chart and Chart editing tasks.} We present the output results of LLaVA-1.5 and ChartLLaMA for the same chart given different instructions. The instruction in the first row requires the model to output the original chart, performing the chart-to-chart task. The instruction in the second row requires the model to output a horizontal bar chart, performing the chart editing task.}
    \label{fig:chart_generation_visualization}
\end{figure*}
\begin{figure*}[t]
    \centering
    \includegraphics[width=0.9\textwidth]{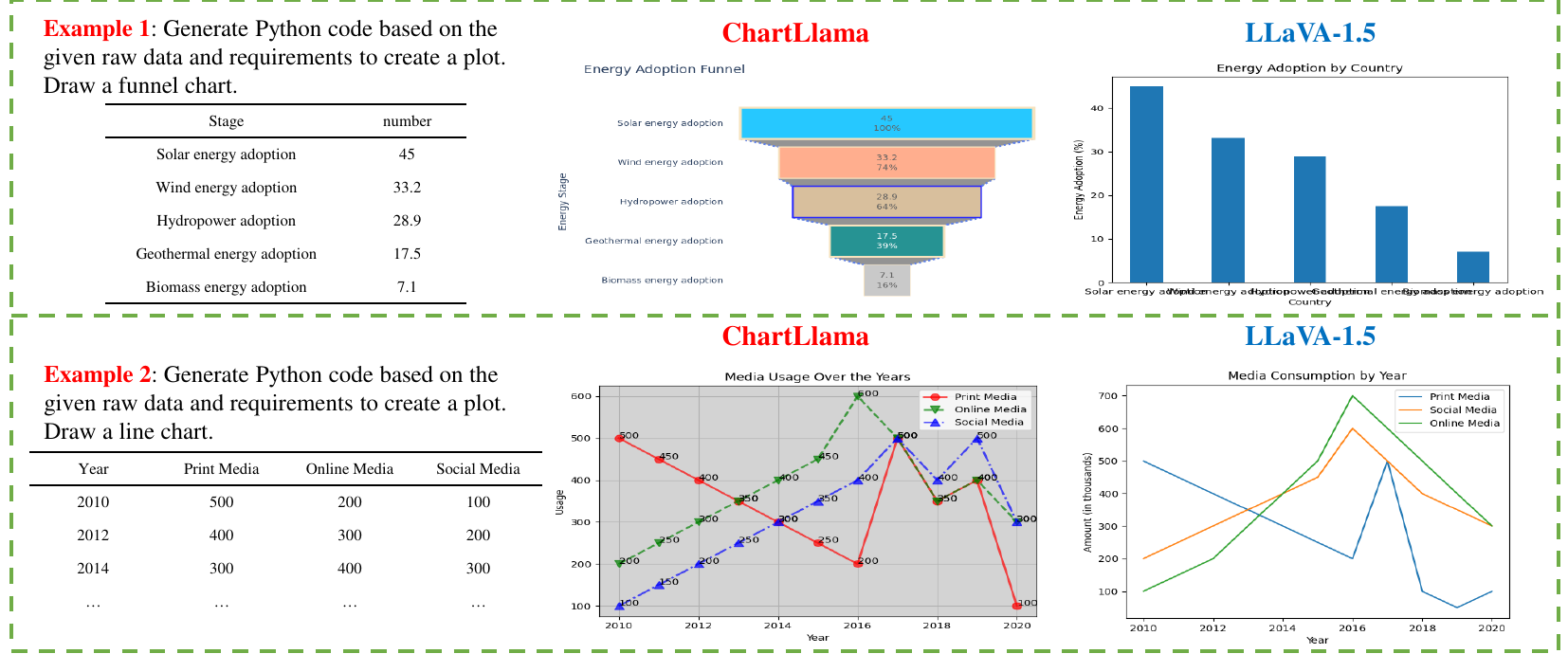}
    \caption{\textbf{Qualitative comparison for Text-to-chart task.} We have presented the generated images by ChartLLaMA and LLaVA-1.5 given the tabular data and the specified requirements. }
    \label{fig:text_to_chart_visualization}
\end{figure*}

\subsection{Results}

We first compare our methods with existing chart understanding models, such as Pix2Struct~\cite{lee2023pix2struct}, Matcha~\cite{matcha}, unichart~\cite{unichart}.
Then we further construct Baseline* using the same model architecture~\cite{llava1.5} as ours, but is trained on the training split of each dataset separately. On traditional tasks, we have also tried to compare with existing multimodal large language models such as InternLM-XComposer~\cite{internlmxcomposer}, MiniGPT-v2~\cite{minigptv2}, and vanilla LLaVA~\cite{llava1.5}. However, we found the limitation of their instruction-following ability makes it hard to be evaluated by existing metrics. 

\noindent\textbf{ChartQA.}
ChartLlama achieves the best performance on both human and augmented splits of ChartQA~\cite{chartqa} as listed in Table~\ref{tab:main_result}.
Previous methods typically involved pretraining on larger datasets and then finetuning on the training split of the same datasets to achieve better results, while ChartLlama does it in a zero-shot way after training on our dataset. 
Notably, although previous methods are trained on the ChartQA's training split, our method achieves significant advantages using much less data as shown in Table~\ref{tab:dataset_statistics}. 
Besides, we also evaluate our model on charts of novel types as shown in Table~\ref{tab:special_chart_results}. Our model gains significant improvement towards Unichart and the Baseline*. This shows the superiority of ChartLlama in the ability to understand charts in novel charts.

\noindent\textbf{Chart-to-text.}
As shown in Table~\ref{tab:main_result} and Table~\ref{tab:gpt_eval}, our method consistently outperforms the previous state-of-the-art approaches under different evaluation metrics and splits in Chart-to-text~\cite{chart-to-text}.
The improvement in our performance primarily stems from the model's ability to handle long texts. Previous works often encountered meaningless repetitions at the end of sentences when dealing with relatively longer texts.

\noindent\textbf{Chart extraction.}
Our model performed the best in this task on ChartQA~\cite{chartqa} as listed in Table~\ref{tab:main_result}.
ChartLlama has been trained on a variety of instruction-tuning data, which greatly improved its ability to understand chart figures. This is the reason why it can significantly outperform LLaVA-1.5 in terms of performance.

\noindent\textbf{Detailed description.}
ChartLlama gains significant performance improvement over LLaVA-1.5 which is shown in Table~\ref{tab:gpt_eval}.
The detailed description task requires the model's ability to understand image details, which can be significantly improved during the training for tasks related to chart figures.

\noindent\textbf{Chart generation and modification.}
In Table~\ref{tab:gpt_eval}, we compare our method with the original LLaVA-1.5, and we can see that our model gains consistent improvement over three tasks. 
LLaVA-1.5, which is the base model of ChartLlama, processes strong abilities to follow instructions and generate Python code, and thus also gains reasonable performances on chart generation and modification tasks.

\subsection{Qualitative results}

Figure~\ref{fig:chart_generation_visualization} visualizes the chart-to-chart and chart-editing results of ChartLlama and our baseline model LLaVA-1.5. ChartLlama plots with the correct color and chart type, while LLaVA-1.5 cannot guarantee the correctness of color, data value, or chart type. 
Figure~\ref{fig:text_to_chart_visualization} shows the text-to-chart results of ChartLlama and LLaVA-1.5. 
In the first example, ChartLlama successfully generates a funnel chart following the instructions and plots correct values. But LLaVA-1.5 even cannot draw funnel charts. In the second example, it is obvious that the result of ChartLlama contains more details and adds data values for human convenience. Both two examples show the strong ability of chart-generating and editing abilities of ChartLlama. 
Our diverse dataset and rich instruction-tuning data have endowed our model with a wide range of practical capabilities.

\section{Conclusion}
In this paper, we propose a flexible and robust approach for synthesizing chart images and instruction-tuning data and train a multimodal LLM on the proposed dataset. Our synthesis process consists of three steps: chart data generation, chart figure generation, and instruction data generation. The data generation flow we propose greatly reduces the difficulty of generating chart-related data for models and improves the controllability and diversity of the generated data.
Experiments conducted on both traditional datasets and our newly constructed dataset validate the outstanding performance of the multimodal LLM. Thanks to the diverse instruction-tuning data in our dataset, the trained multimodal language model possesses various capabilities that were absent in previous models. Moreover, its ability to comprehend both instructions and figures can easily extend to new categories of chart figures or tasks.
We believe that our data generation process can make significant contributions to multimodal LLM in tasks related to chart understanding. Furthermore, it will facilitate the application of similar data generation processes in other domains.

\noindent\textbf{Limitations.} The current version of ChartLlama's vision encoder lacks the ability to handle multilingual OCR tasks, restricting the model's utility for charts containing non-English text. To overcome this limitation, we are contemplating the creation of a novel vision encoder that boasts proficiency in multilingual OCR tasks.
{
    \small
    \bibliographystyle{ieeenat_fullname}
    \bibliography{main}
}

\appendix
\twocolumn[{
\renewcommand\twocolumn[1][]{#1}
% \maketitle
\maketitlesupplementary
\centering
\vspace{-0.5cm}
    \resizebox{\textwidth}{!}{
        \begin{tabular}{l|ccc|cccccc} \Xcline{1-10}{1pt}
\multirow{2}{*}{Method} & \multicolumn{3}{c}{ChartQA} & \multicolumn{6}{c}{ChartQA on special charts}  \\ \Xcline{2-10}{1pt}
 & Human & Augmented & Average & Funnel & Gantt & Heatmap & Scatter & Box & Candlestick \\ \Xcline{1-10}{1pt}
Unichart~\cite{unichart} & 43.92 & 88.56 & 66.24 & 18.30 & 9.80 & 25.43 & 26.32 & 16.67 & 15.79 \\
InternLM-XComposer-VL~\cite{internlmxcomposer} &8.48 & 7.36 & 7.92 & 12.42 & 6.36 & 16.18 & 18.13 & 15.33 & 16.96 \\
Mini-GPT-v2~\cite{minigptv2} & 15.60 & 8.40 & 12.00 & 26.7 & 15.03 & 28.32 & 28.65 & 21.33 & 17.54 \\
Qwen-VL~\cite{Qwen-VL} & 37.60 & 63.76 & 50.68 & 6.54 & 9.83 & 13.29 & 7.02 & 8.00 & 1.75 \\
mPLUG-Owl2~\cite{ye2023mplugowl} & 21.20 & 22.0 & 21.60 & 23.53 & 27.75 & 19.08 & 16.37 & 15.33 & 19.30 \\
Baseline*~\cite{llava1.5} & 37.68 & 72.96 & 55.32 & 49.32 & 40.17 & 38.18 & 37.91 & 28.33 & 25.69 \\
\textbf{ChartLlama} & \textbf{48.96} & \textbf{90.36} & \textbf{69.66} & \textbf{70.59} & \textbf{56.64} & \textbf{53.18} & \textbf{54.97} & \textbf{37.33} & \textbf{46.20} \\  \Xcline{1-10}{1pt}
\end{tabular}
    }
    \vspace{-0.1in}
    \captionsetup{type=table}
\caption{\textbf{Results on traditional tasks.} We compare our work with the previous three open-source models and also compare it with Baseline* trained on the training split of respective benchmarks. }
    \label{tab:appendix_main_result}
\vspace{0.2cm}
}]
\setcounter{page}{1}
% \maketitlesupplementary

\begin{figure}[t]
    \centering
    \includegraphics[width=0.45\textwidth]{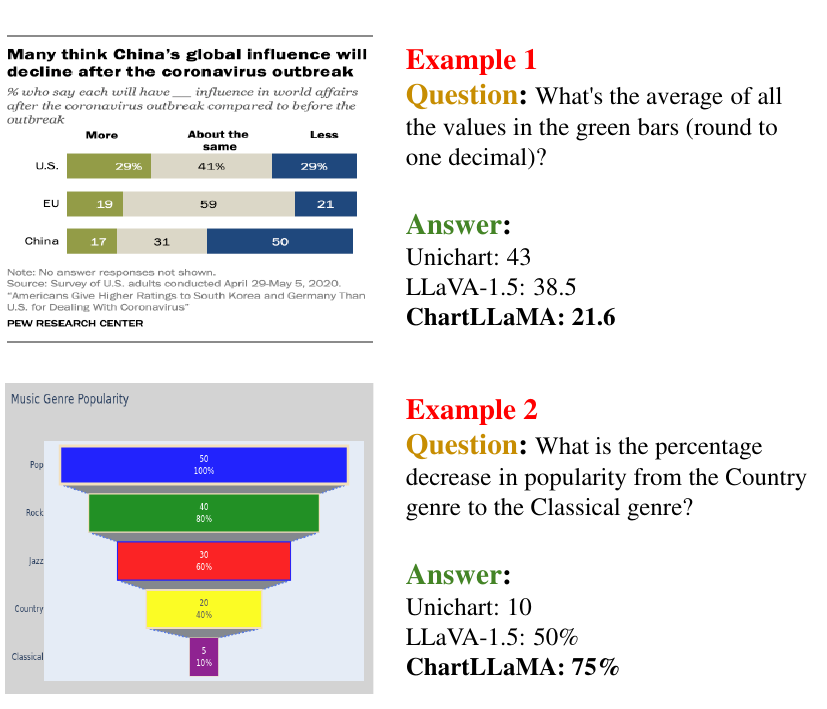}
    \caption{\textbf{Visualization on the ChartQA task.} Here are two examples of the predictions of Unichart, LLaVA-1.5, and ChartLlama. Our proposed ChartLlama could follow the long instructions and do calculations to get the correct results.}
    \label{fig:chartqa_visualization}
\end{figure}
\section{Model architecture}\label{appendix:model_architecture}

To elucidate our training strategies, we provide some clarification about the modifications in LLaVA-1.5~\cite{llava1.5}, and introduce its essential model architectures.

\paragraph{Vision encoder:} LLaVA-1.5 incorporates CLIP's vision encoder~\cite{clip}. The primary distinction is that LLaVA-1.5 employs ViT-L/14@336px, while LLaVA uses ViT-L/14@224px. Another notable alteration concerns the image processor. Eschewing traditional center cropping, LLaVA-1.5 adopts padding as an image pre-processing technique, ensuring that all information in the provided image can be apprehended.

\paragraph{Projection layer:} In LLaVA-1.5, the initial single linear layer is substituted with a two-layer MLP, resulting in improved performance.

\paragraph{Lora Layer:} Based on experiments in~\cite{scalellava,llava1.5}, implementing Lora~\cite{lora} layers is sufficient to achieve performance comparable to full fine-tuning strategies. For the original LLaVA~\cite{llava}, Lora layers with a Lora rank of 64 suffice, whereas for LLaVA-1.5~\cite{llava1.5}, the Lora rank needs to exceed 128.

\begin{table}[t]
\begin{tabular}{lc}\toprule
\makebox[0.12\textwidth][l]{Prompt Design} & \makebox[0.22\textwidth][c]{Successful Rate} \\ \midrule
Original & 85\%\\ 
\textit{w/o} In context example & 43\%\\
\textit{w/o} Documentation & 65\%\\
\textit{w/o} Both & 28\%\\ \bottomrule
\end{tabular}
\caption{\textbf{Ablations on Prompt of Stage Two.} The first row shows the successful rate of our proposed data generation method in the second stage. Then we evaluate the generated results when removing the in-context examples, the documentation, and both of them, respectively.}
\label{tab:stage2_ablations}
\end{table}

\begin{figure*}[t]
    \centering
    \includegraphics[width=0.95\textwidth]{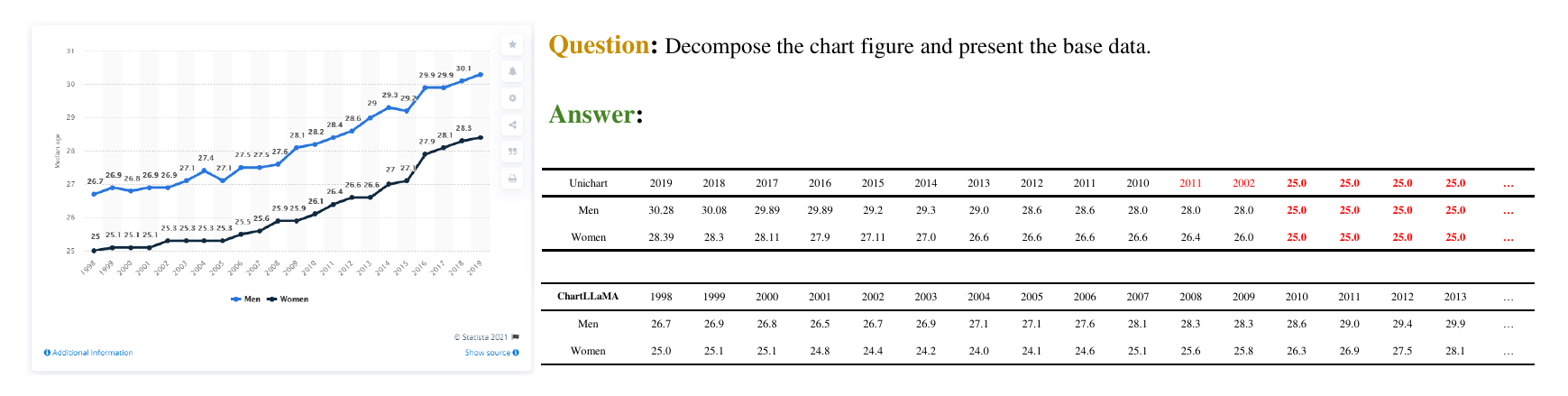}
    \caption{\textbf{Visualization of Chart-extraction.} We find that ChartLlama is especially good at long text processing. While the previous SOTA, Unichart, will generate meaningless redundant words when the output is too long.}
    \label{fig:chart_extraction_visualization}
\end{figure*}
\begin{figure*}[t]
    \centering
    \includegraphics[width=0.95\textwidth]{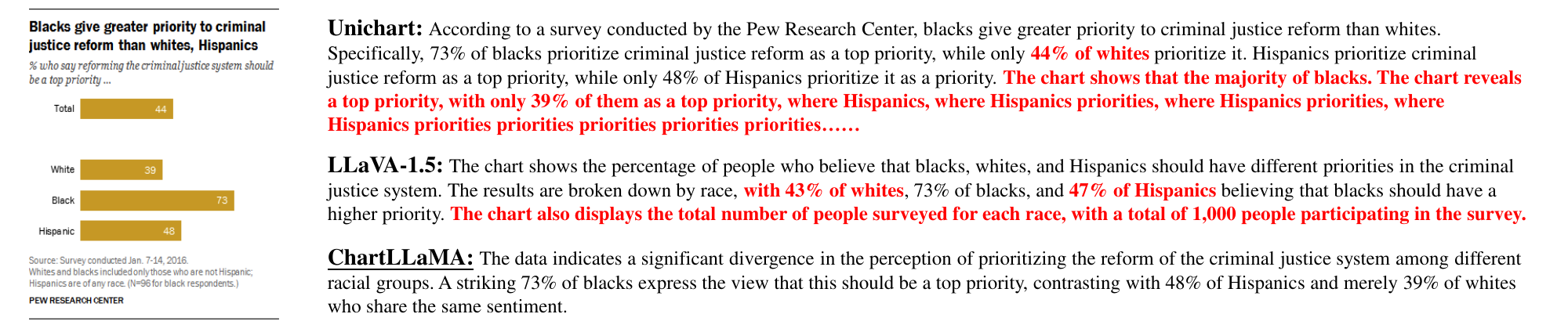}
    \caption{\textbf{Visualization of Chart-to-text.} We select one image from the Pew Dataset and show the results of Unichart, LLaVA-1.5, and ChartLlama. We find that Unichart easily falls into repeated words again and LLaVA-1.5 suffers from hallucination.}
    \label{fig:chart_to_text_visualization}
\end{figure*}

\begin{figure*}[t]
    \centering
    \includegraphics[width=0.95\textwidth]{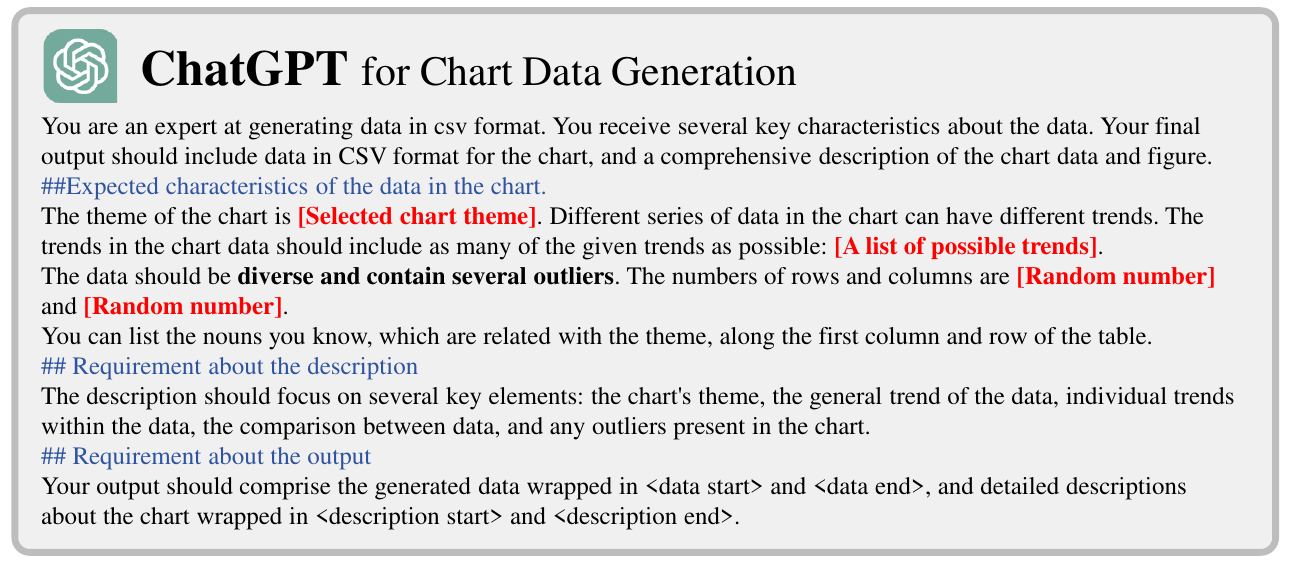}
    \caption{\textbf{The prompt template for Stage One.} This template is used for Chart Data Generation. Utilizing this template could guide GPT-4 to generate diverse raw tabular data and detailed descriptions of the content.}
    \label{fig:stage1}
\end{figure*}
\begin{figure*}[t]
    \centering
    \includegraphics[width=0.95\textwidth]{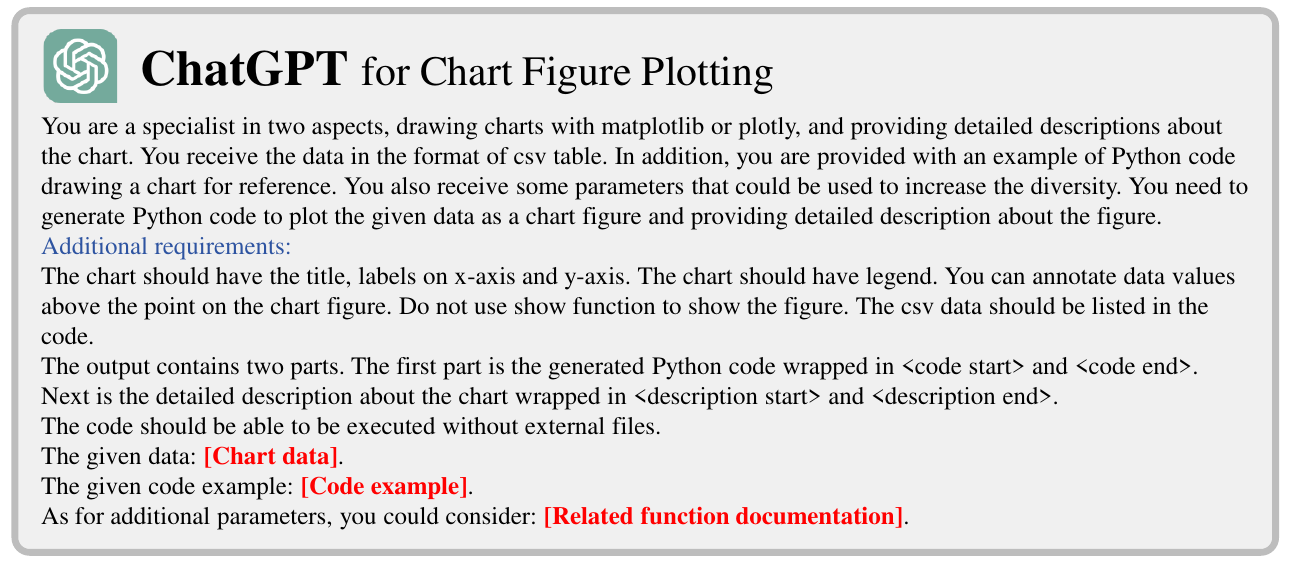}
    \caption{\textbf{The prompt template for Chart Figure Plotting.} Following such instructions, GPT-4 could generate codes that could draw chart figures using Python packages.}
    \label{fig:stage2}
\end{figure*}
\begin{figure*}[t]
    \centering
    \includegraphics[width=0.95\textwidth]{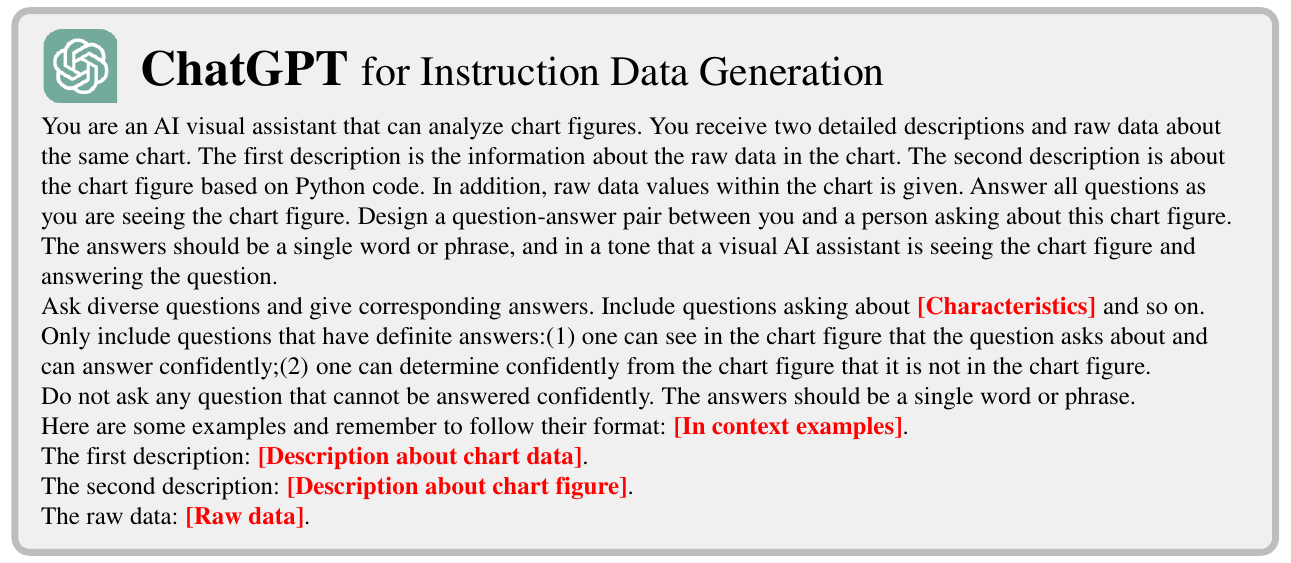}
    \caption{\textbf{The prompt template for Instruction Data Generation.} This step is targeted at generating various training data. To guarantee the quality and diversity of the generated samples, we need to give enough information on the chart figure and in-context examples.}
    \label{fig:stage3}
\end{figure*}
\begin{figure*}[t]
    \centering
    \includegraphics[width=0.95\textwidth]{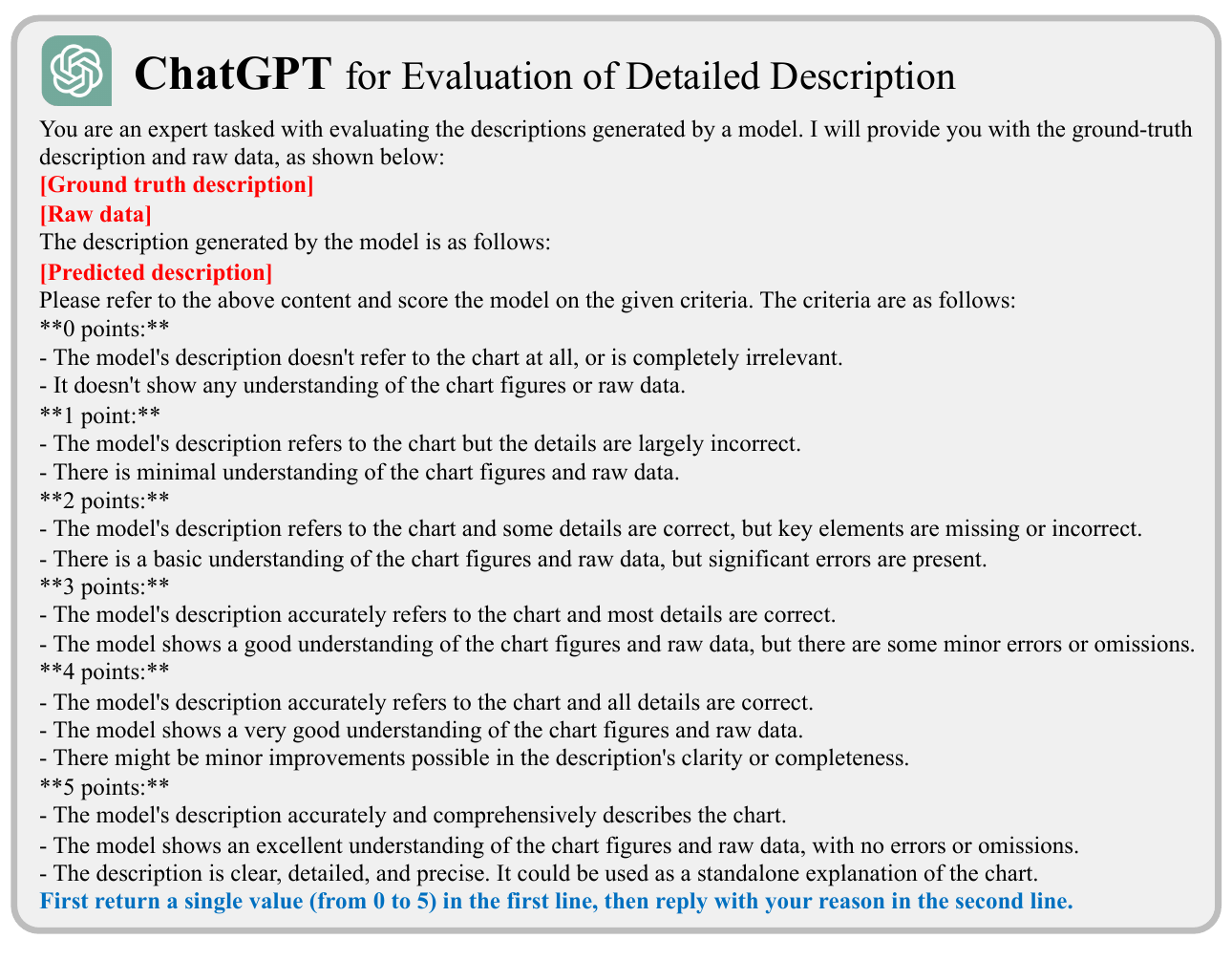}
    \caption{\textbf{The prompt template used for evaluation on the Detailed-description task.} The input conditions are the ground-truth description, raw data, and predicted description. GPT-4 will follow the given criteria and generate the final score and reasons.}
    \label{fig:detailed_description}
\end{figure*}
\begin{figure*}[t]
    \centering
    \includegraphics[width=0.95\textwidth]{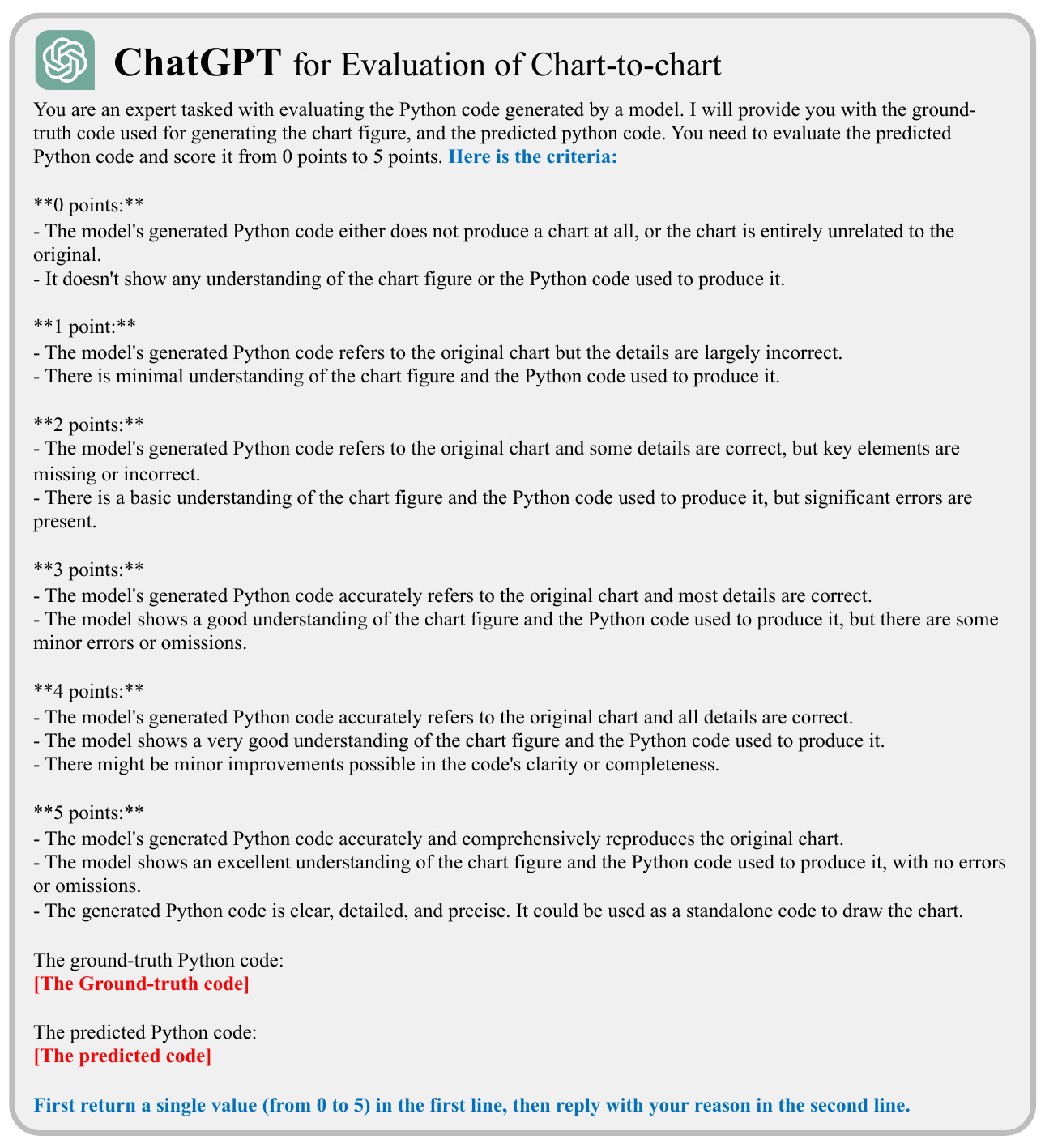}
    \caption{\textbf{The prompt template used for evaluation on the Chart-to-chart task.} The input conditions are ground-truth code and predicted code. Following the given criteria, GPT-4 generates the score and corresponding reason. }
    \label{fig:chart_to_chart}
\end{figure*}
\begin{figure*}[t]
    \centering
    \includegraphics[width=0.95\textwidth]{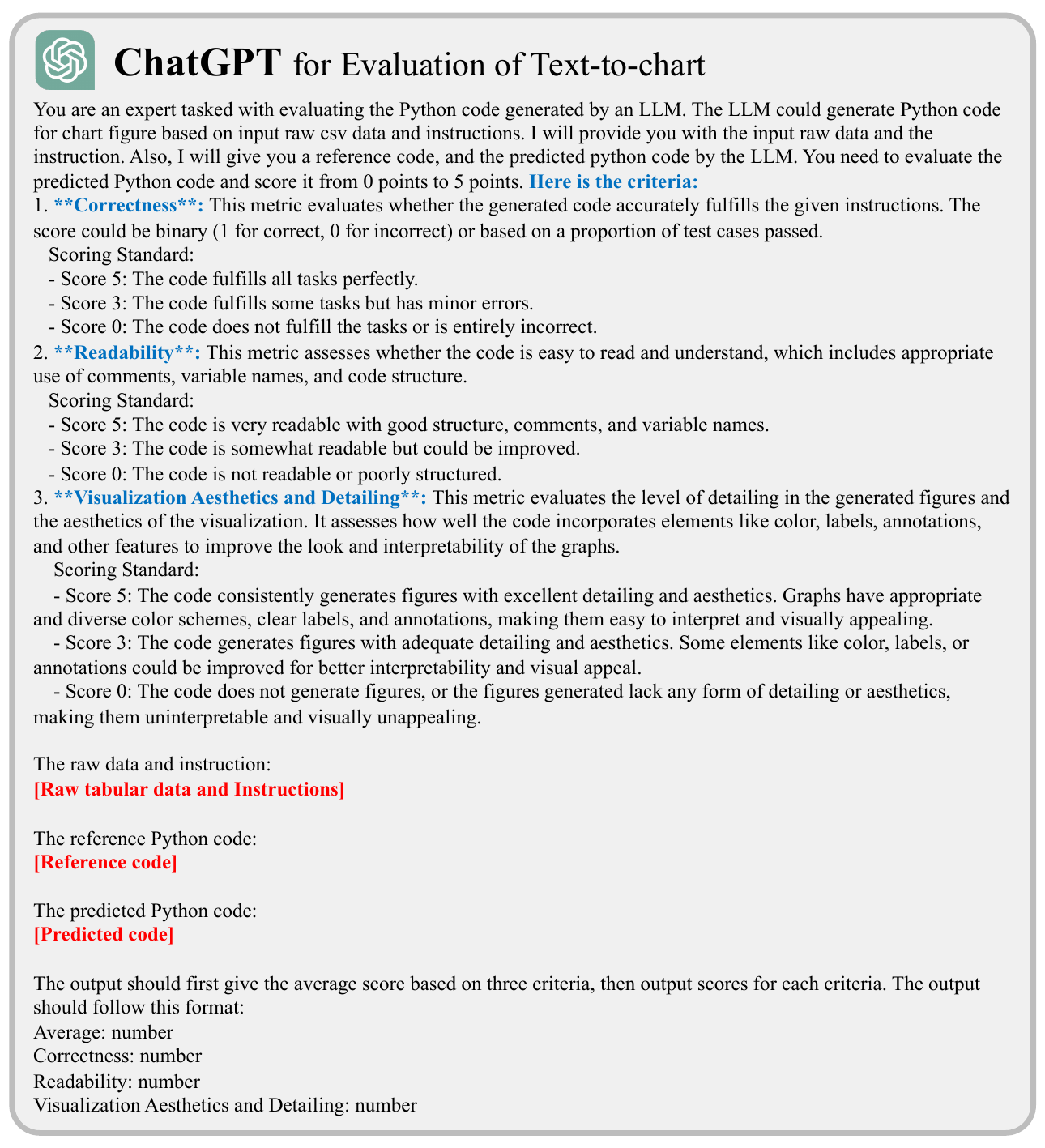}
    \caption{\textbf{The prompt template used for evaluation on the Text-to-chart task.} The input conditions are the raw tabular data and instructions, the reference code, and the predicted code. Finally, GPT-4 will return with the average score and scores for each criterion.}
    \label{fig:text_to_chart}
\end{figure*}
\begin{figure*}[t]
    \centering
    \includegraphics[width=0.95\textwidth]{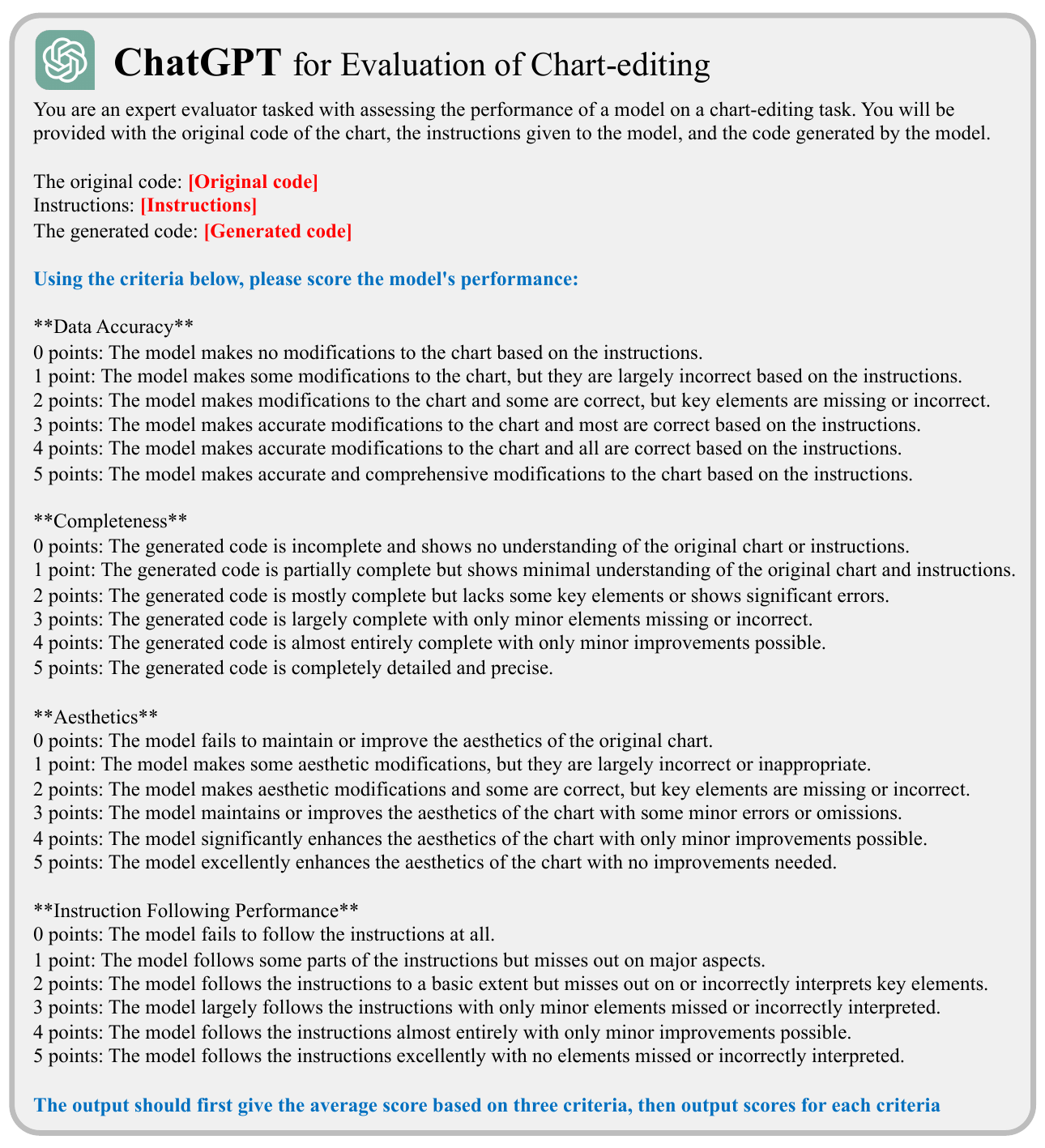}
    \caption{\textbf{The prompt template used for evaluation on the Chart-editing task.} Input conditions include the original code corresponding to the given chart figure, the instructions that describe how to edit the figure, and the generated code. The output will contain the final average score and scores for each criterion. }
    \label{fig:chart_editing}
\end{figure*}
\begin{figure*}[t]
    \centering
    \includegraphics[width=0.95\textwidth]{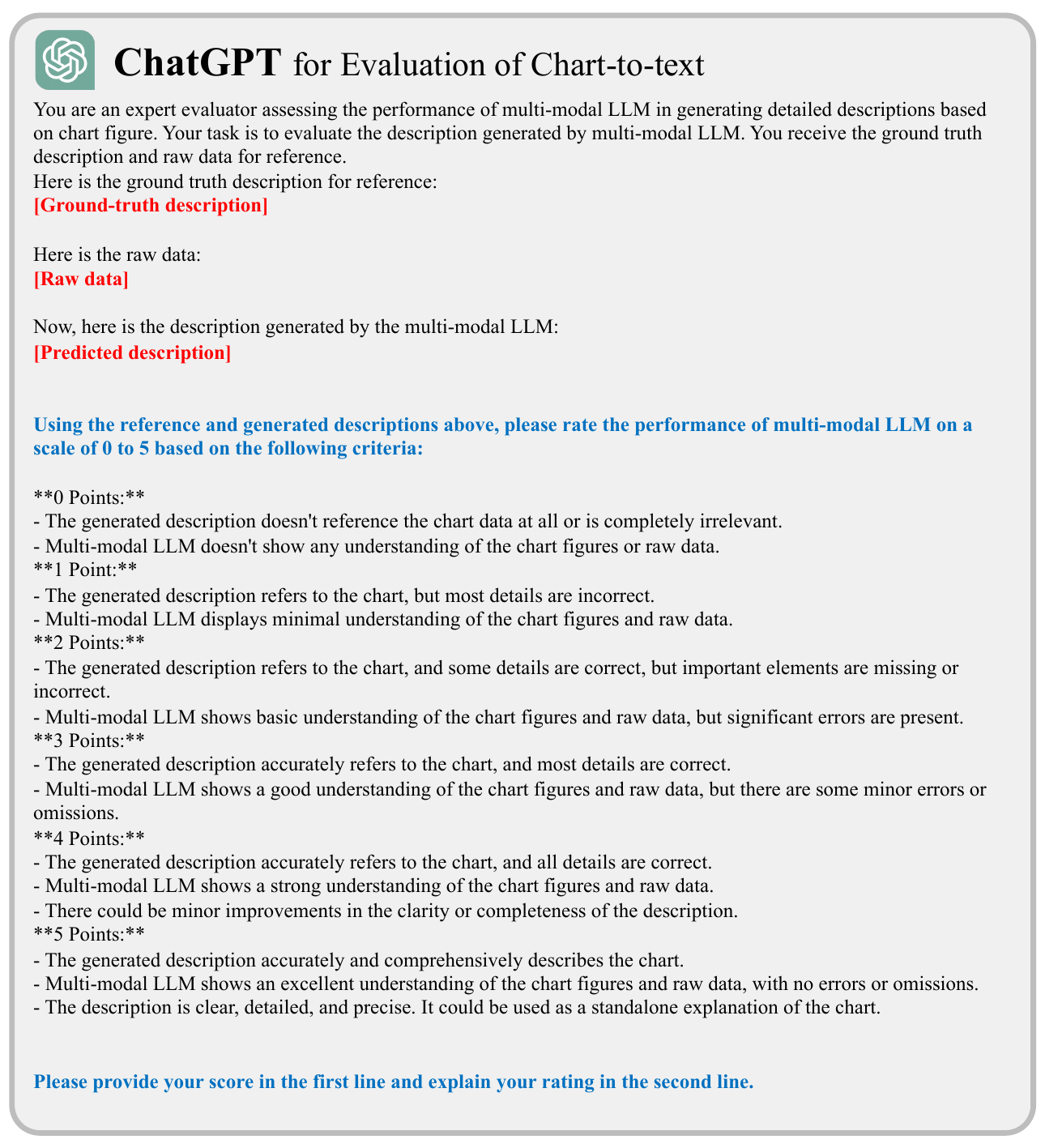}
    \caption{\textbf{The prompt template used for evaluation on the Chart-to-text task.} The ground-truth description, raw data, and predicted description are input conditions. This evaluation prompt requires GPT-4 to give the final score and explanations of the given score.}
    \label{fig:chart_to_text}
\end{figure*}

\section{Dataset Scale}

The model's training process is broken down into two critical stages: pretraining and fine-tuning. The primary objective of pretraining is to effectively initialize the vision projector while fine-tuning steers the Language Learning Model (LLM) to adhere to the provided instructions. 

In the pretraining phase, LLaVA-1.5 utilizes approximately 558k image-caption pairs to train the projection layer. It is anticipated that the vision features will align with the language features to a certain extent. This dataset originates from a subset of around 558K image-text pairs from LAION-CC-SBU, each paired with a BLIP caption.

The fine-tuning phase involves further training of the model on 665k instruction-following data pairs. LLaVA-1.5 manifests an array of capabilities during this stage. The instruction-following data pairs are meticulously generated to encompass the required abilities. To enhance the model's capacities in varied contexts, additional academic-task-focused Visual Question-Answering (VQA) datasets for VQA, Optical Character Recognition (OCR), and region-level perception are incorporated. The final compilation includes several datasets: OpenKnowledge VQA (OKVQA, A-OKVQA), Region-level VQA (Visual Genome, RefCOCO), and OCR (OCRVQA, TextCaps). A-OKVQA is transformed into multiple-choice questions, employing a specific response formatting prompt: answer by directly specifying the option’s letter from the provided choices.

\section{Generation Prompt for ChartLlama}
As listed in Figure~\ref{fig:stage1}, Figure~\ref{fig:stage2}, and Figure~\ref{fig:stage3}, we have provided standard prompts for data generation in three stages. The text in black color in the figure denotes the fixed prompt template, while the text in red color brackets requires filling in, which serves to enhance the diversity and controllability of the generated results. The detailed meanings of the different variables have already been discussed in the main text, thus we will not elaborate further.

\section{Ablation Study on the Conditions of Generation Prompt for ChartLlama}
In order to verify the impact of our proposed generation process on the results, we designed an ablation experiment on the prompt for the second step, diagram construction, which is shown in Table~\ref{tab:stage2_ablations}. Specifically, we removed the in-context examples and the description of the function, then retested the probability of successful generation. The results show that combining both in-context examples and documentation could significantly improve the successful rate of plotting figures. Also, we observe that the diversity could also improve a lot, which is hard to quantify. 

\section{Filtering Mechanism}
The data generation process may produce some erroneous samples, but filtering and correcting these samples can be challenging because the samples contain figures that cannot be processed by GPT-4. We only performed basic error correction, including checking the data generation format and verifying the correct execution of the code. The data generation format check involves confirming whether the model has separated different data results with different markers according to our requirements. The check for correct code execution involves running the generated plotting script. If this script fails to run, we no longer use the training sample corresponding to that plot. Such basic data screening is sufficient to ensure the quality of the generated dataset. We are also considering incorporating more effective automatic screening mechanisms to avoid contamination of the dataset by poor-quality samples.

\section{Evaluation Prompt for ChartLlama}
We have prepared five evaluation prompts in total, each tailored for a specific task: chart-to-text in Figure~\ref{fig:chart_to_text}, detailed description in Figure~\ref{fig:detailed_description}, chart-to-chart in Figure~\ref{fig:chart_to_chart}, text-to-chart in Figure~\ref{fig:text_to_chart}, and chart-editing in Figure~\ref{fig:chart_editing}. We have designed distinctive scoring criteria for different tasks and provided reference information based on the additional annotations in the dataset. Ultimately, we employed GPT-4 for scoring purposes.

\section{Comparison with Multi-modal LLMs}

\paragraph{Traditional Tasks:} The Table~\ref{tab:appendix_main_result} includes a comparison of existing state-of-the-art (SOTA) models, illustrating their respective performances. Interestingly, some models~\cite{internlmxcomposer} show unexpectedly low performance. This outcome is not a consequence of our experimental configuration. Rather, it derives from the fact that these models have not been trained on corresponding instruction-following tasks, which results in outputs that are incompatible with the evaluation framework. We argue that training these models specifically on instruction-following tasks using specific datasets would likely yield improved performance. Another notable observation is the performance gap of Qwen-VL between the ChartQA test splits and the ChartQA on our specially generated charts. Despite being trained on ChartQA, Qwen-VL underperforms on the specially generated charts, underscoring the effectiveness and need for our proposed benchmark. However, the lack of general training scripts provided by many models poses a challenge to our fine-tuning efforts. Nonetheless, our hypothesis finds support in the model LLaVA-1.5. Initially, LLaVA-1.5 performed poorly on the dataset but showed significant improvement when trained on the designated dataset.

\paragraph{Novel Tasks:}
We also conducted tests on the newly proposed tasks. However, most of the given dataset cannot generate executable Python code except LLaVA-1.5~\cite{llava1.5}. We speculate that this is because these large multimodal models have been overtrained on visual language datasets, resulting in the loss of their code generation capabilities in Language Learning Models (LLMs); while LLaVA-1.5 adopted a series of optimization measures during its training process. For instance, compared to other large multimodal language models, LLaVA-1.5 has a shorter training time, fewer training parameters, a more moderate dataset scale, and incorporates pure text data during training to maintain the basic capabilities of LLMs. This experiment also suggests that if we expect the model to have a certain level of generalization ability, we should avoid making excessive adjustments to the LLMs. This is also why our ChartLlama model chose to train with fewer parameters.

\section{More Qualitative Results}
\paragraph{ChartQA.}
As shown in Figure~\ref{fig:chartqa_visualization}, we compare our ChartLLaMA with Unichart and LLaVA-1.5. The given examples are both related to longer questions and calculations, which is hard for Unichart. 
What's more, without the language understanding ability, Unichart even cannot follow complex instructions. In Example 2, the answer of Unichart is even not a percentage. 
Although LLaVA-1.5 has the ability of OCR and instruction-tuning, it cannot identify which part of the image is related to the question because it has not been trained on chart figures. Thus, it fails in both examples, either.

\paragraph{Chart Extraction.}
As depicted in Figure~\ref{fig:chart_extraction_visualization}, ChartLLaMA also possesses the capability to convert charts into structured data. Both the output results of Unichart and ChartLLaMA are a string of characters and we visualize it as tables for convenience. The first mistake of Unichart is reversing the order of years. Another mistake in Unichart is the persistent output of repetitive and meaningless characters at the end. Meanwhile, our proposed model, ChartLLaMA, benefits from strong language comprehension and output capabilities, which prevent the occurrence of such errors.

\paragraph{Chart Description.}
In Figure~\ref{fig:chart_to_text_visualization}, we visualize the results of Unichart, LLaVA-1.5, and ChartLLaMA on the Chart-to-text task. The results from Unichart contain incorrect values and meaningless repetitions when generating long texts. LLaVA-1.5 performs better for long output sequences due to the strong language understanding and generation abilities of the LLM backbone. However, it suffers from wrong OCR recognition results and hallucinations. Our proposed ChartLLaMA performs best among these three models. 

\end{document}